\documentclass{article}

\usepackage{microtype}
\usepackage{graphicx}
\usepackage{subfigure}
\usepackage{booktabs} 
\usepackage{makecell}
\usepackage{hyperref}

\usepackage[accepted]{icml2024}

\usepackage{amsmath}
\usepackage{amssymb}
\usepackage{mathtools}
\usepackage{amsthm}
\usepackage{multirow}
\usepackage{graphicx}
\usepackage{bbding}
\usepackage[capitalize,noabbrev]{cleveref}

\theoremstyle{plain}

\theoremstyle{definition}

\theoremstyle{remark}

\usepackage[textsize=tiny]{todonotes}

\icmltitlerunning{BLO-SAM: Bi-level Optimization Based Overfitting-Preventing  Finetuning of SAM}

\begin{document}

\twocolumn[
\icmltitle{BLO-SAM: Bi-Level Optimization Based Finetuning of the Segment Anything Model for Overfitting-Preventing Semantic Segmentation}




\begin{icmlauthorlist}
\icmlauthor{Li Zhang}{1}
\icmlauthor{Youwei Liang}{1}
\icmlauthor{Ruiyi Zhang}{1}
\icmlauthor{Amirhosein Javadi}{1}
\icmlauthor{Pengtao Xie}{1}
\end{icmlauthorlist}

\icmlaffiliation{1}{University of California, San Diego}

\icmlcorrespondingauthor{Pengtao Xie}{}

\icmlkeywords{Semantic Segmentation, End-to-End finetune, Segment Anything Model}

\vskip 0.3in
]



\printAffiliationsAndNotice{}  

\begin{abstract}

The Segment Anything Model (SAM), a foundation model pretrained on millions of images and segmentation masks, has significantly advanced semantic segmentation, a fundamental task in computer vision. 
Despite its strengths, SAM encounters two major challenges. Firstly, it struggles with segmenting specific objects autonomously, as it relies on users to manually input prompts like points or bounding boxes to identify targeted objects. 
Secondly, SAM faces challenges in excelling at specific downstream tasks, like medical imaging, due to a disparity between the distribution of its pretraining data, which predominantly consists of general-domain images, and the data used in downstream tasks. Current solutions to these problems, which involve finetuning SAM, often lead to overfitting, a notable issue in scenarios with very limited data, like in medical imaging.
To overcome these limitations, we introduce BLO-SAM, which finetunes  SAM based on bi-level optimization (BLO). Our approach allows for automatic image segmentation without the need for manual prompts, by optimizing a learnable prompt embedding.  Furthermore, it significantly reduces the risk of overfitting by training the model's weight parameters and the prompt embedding on two separate subsets of the training dataset, each at a different level of optimization. We apply BLO-SAM to diverse semantic segmentation tasks in general and medical domains. The results demonstrate BLO-SAM's superior performance over various state-of-the-art image semantic segmentation methods. The code of BLO-SAM is available at \url{https://github.com/importZL/BLO-SAM}.
\end{abstract}

\section{Introduction}\label{introduction}
Semantic segmentation is a critical task in computer vision, which aims to assign each pixel with a semantic class (object classes such as dog and cat, or scene categories such as sky and ocean)~\cite{mo2022review}. 
Deep learning has demonstrated great success in advancing the performance of semantic segmentation~\cite{lateef2019survey}. 
This progress has been further propelled by the emergence of foundation models (FMs)~\cite{bommasani2021opportunities}, a.k.a. large pretrained models, which have demonstrated unprecedented prevalence across diverse tasks, including  vision~\cite{clip,moor2023foundation}, language~\cite{brown2020language,touvron2023llama,Touvron2023llama2}, and multi-modality~\cite{alayrac2022flamingo,wang2022simvlm}. 
Building on a data engine using 11 million image-mask pairs, the Segment Anything Model (SAM)~\cite{kirillov2023segment} emerges as a noteworthy segmentation foundation model, and demonstrates strong capabilities in segmenting diverse natural images. As a novel promptable segmentation model, SAM distinguishes itself by yielding desired segmentation masks when provided with appropriate points or bounding boxes as prompts (detailed descriptions about SAM can be found in Appendix~\ref{sam_pre}).

Despite its strong performance in producing accurate segmentation masks based on prompts, SAM faces a notable limitation - it cannot autonomously segment specific objects, as it requires points or bounding boxes as manual prompts\footnote{Although the SAM paper mentions that SAM can accept textual prompts, only points, bounding boxes, and coarse masks are supported as prompts in its public codebase.}. For example, if we would like to segment lungs from a chest X-ray image, we need to provide at least one point on the lung region or bounding boxes enclosing the lungs to indicate that the objects aimed to segment are the lungs. 
Another significant challenge of applying SAM for downstream segmentation tasks arises from the distribution discrepancy between the pretraining data of SAM and the data in downstream tasks. 
On  tasks where the data distribution deviates from the pretraining data, SAM struggles to segment the desired objects accurately even with proper prompts~\cite{mazurowski2023segment,he2023computer}.

\begin{figure*}
    \centering
    \label{fig:overview}
    \includegraphics[width=1\linewidth]{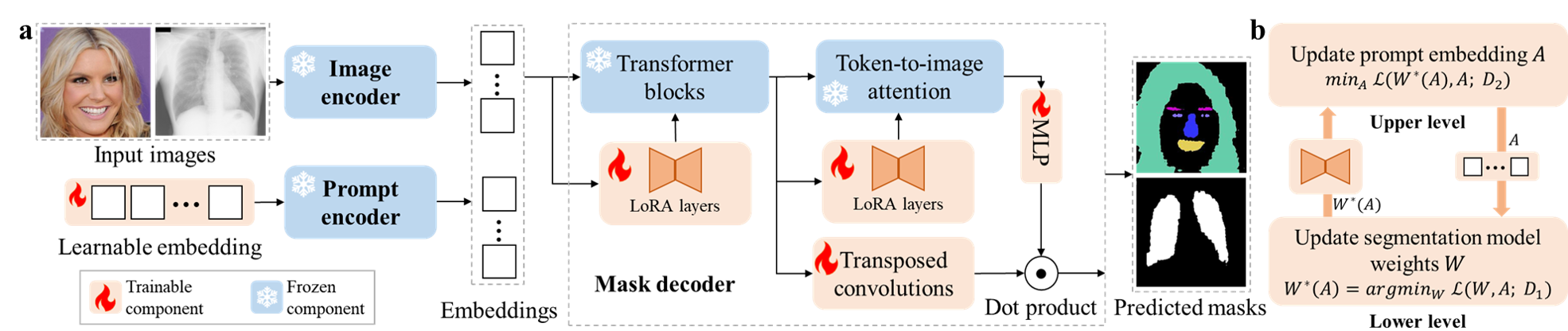}
    \vspace{-0.7cm}
    \caption{\textbf{(a) }BLO-SAM Overview: The majority of original SAM parameters are frozen. LoRA layers enable parameter-efficient finetuning of the mask decoder's Transformer layers.  Learnable parameters are categorized into two groups: learnable model parameters in the mask decoder and learnable embeddings as input to the prompt encoder. \textbf{(b) }Two parameter groups are updated through two interdependent optimization problems.}
    \vspace{-0.5cm}
\end{figure*}

\begin{figure}
    \centering
    \includegraphics[width=1\linewidth]{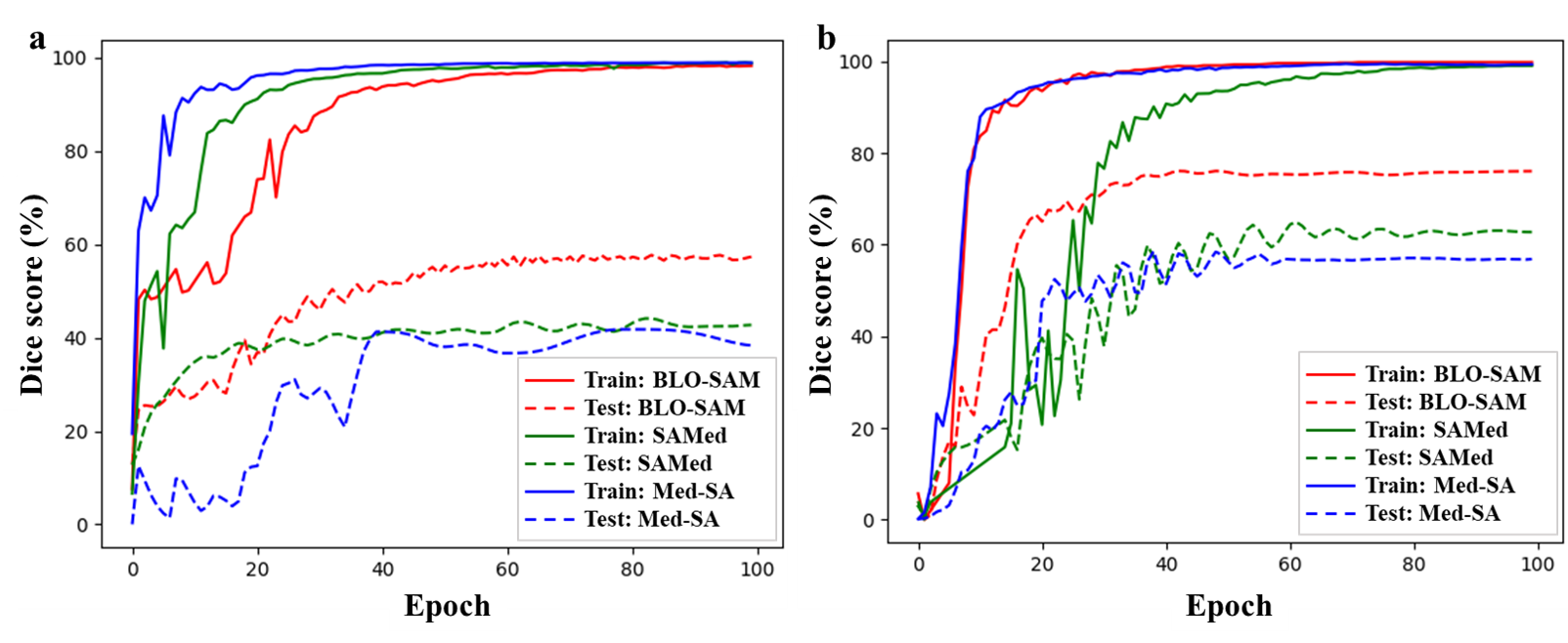}
    \vspace{-0.8cm}
    \caption{Dice scores on the train and test sets versus training epochs, on \textbf{(a) }gastrointestinal disease and \textbf{(b) }human body segmentation tasks. The train and test curves for SAMed exhibit a large gap between epochs 60 and 100, indicating that severe overfitting to the training data occurs. A similar pattern is observed in the case of Med-SA. Conversely, the train-test gap for BLO-SAM is much smaller compared to SAMed and Med-SA, underscoring our method's effectiveness in mitigating overfitting.
 }
    \label{fig:train_process}
    \vspace{-0.5cm}
\end{figure}

To address the distribution discrepancy issue, Med-SA~\cite{wu2023medical} utilizes Adapter~\cite{houlsby2019parameter} to finetune SAM's image encoder and mask decoder in downstream tasks, resulting in superior performance over several SOTA segmentation methods on various medical datasets. Nevertheless, Med-SA still needs prompts manually provided by humans or generated from an initial segmentation mask, during both training and inference phases, making it less practical in real-world applications. 
To address both challenges (i.e., distribution discrepancy and the need for manual prompts), SAMed~\cite{zhang2023customized} finetunes SAM with a  default prompt (i.e., a learnable vector) instead of manual prompts, directly outputting multiple segmentation masks for all the classes. 
However, the above methods still have limitations. They bear a risk of overfitting when labeled data are very limited in downstream tasks. For example, in medical domains, images with labeled segmentation masks are often scarce, either due to the limited number of input images  (e.g., for privacy concerns) or the difficulty of obtaining segmentation masks (e.g., requiring domain experts to annotate, which is time-consuming and expensive). Finetuning large foundation models like SAM on these label-scarce settings often leads to overfitting to training data and poor generalization to test images. 
As demonstrated in our experiments (see Fig.~\ref{fig:train_process}), SAMed and Med-SA suffer from severe overfitting and low test performance.

In response to the above challenges, we introduce BLO-SAM, a finetuning method for SAM that addresses the overfitting issue based on  bi-level optimization (BLO). BLO-SAM combats overfitting by updating two separate sets of learnable parameters on two splits of the training data. Illustrated in Fig.~\ref{fig:overview}(a), BLO-SAM involves two sets of parameters: i) segmentation model's weight parameters, including LoRA layers~\cite{hu2021lora}, transposed convolutions, and MLP heads, and ii) the prompt embedding. These parameters undergo optimization through two levels of nested optimization problems, as shown in Fig.~\ref{fig:overview}(b). 
In the lower level, we iteratively train segmentation model weights by minimizing a finetuning loss on a designated subset of the training dataset while keeping the learnable prompt embedding fixed. Following this finetuning, we transition to the upper level to validate the model's effectiveness on the remaining subset of the training data. The prompt embedding is then updated by minimizing the validation loss. 
By segregating the learning processes for different learnable parameters onto distinct data subsets within two optimization problems, our method effectively mitigates the risk of overfitting to a single dataset, enhancing the model's generalization on test examples. 

BLO-SAM's mechanism of combating overfitting is inspired by the well-established practice of hyper-parameter tuning~\cite{franceschi2018bilevel}. Typically, a model’s weight parameters (such as weights and biases in a neural network) are trained on a training dataset, while the hyper-parameters (like the number of layers in a neural network) are tuned on a separate validation set, which prevents overfitting the training data. If hyper-parameters were also tuned on the training set, it would lead to significant overfitting. Similarly, in the context of SAM, prompt embeddings can be regarded as a form of `hyper-parameters'. Overfitting arises when these embeddings, along with the segmentation model weights, are optimized together by minimizing a loss function on a single dataset, as SAMed do. Our BLO-SAM follows the correct way of `hyper-parameter tuning': it optimizes the `hyper-parameters' - the prompt embedding - on a `validation set' - a subset of the training data, while training the segmentation model weights on the other subset. 




Our work makes two key contributions:
\begin{itemize}
    \vspace{-0.2cm}
    \item We propose BLO-SAM, an overfitting-resilient approach for finetuning SAM with only a few training examples. We propose to learn the prompt embedding and model parameters of SAM on two distinct sub-datasets in a BLO framework, effectively combating overfitting in ultra-low data regimes. Furthermore, our method enables fully automated segmentation without the need for manual prompts during inference and training. This substantially enhances the practical applicability of SAM in real-world scenarios.  
    \vspace{-0.1cm}
    \item Our extensive experiments on six datasets from general domains and medical domains demonstrate the strong effectiveness of BLO-SAM with less than 10 training examples. Our method significantly outperforms vanilla SAM, SAM-based models, few-shot semantic segmentation methods, and popular segmentation models, without requiring manual prompts.  
\end{itemize}

\section{Related Work}\label{related}

\subsection{Semantic Segmentation}
In the realm of semantic segmentation, substantial progress has been made  recently, particularly within the context of deep learning-based methods. Many works focus on the architectural design of segmentation neural networks, such as fully convolutional networks (FCNs)~\cite{long2015fully}, U-Net~\cite{ronneberger2015u},  DeepLab~\cite{chen2017deeplab}, and SegFormer~\cite{xie2021segformer}. The incorporation of attention mechanisms, exemplified in Atrous Spatial Pyramid Pooling (ASPP)~\cite{chen2017deeplab} and non-local neural networks~\cite{wang2018non}, has further improved contextual understanding in semantic segmentation. Recently, the Segment Anything Model (SAM)~\cite{kirillov2023segment} emerged as an FM for image segmentation. SAM uses an MAE-pretrained ViT~\cite{he2022masked} to encode the input image, positional embedding to encode prompts, and a lightweight Transformer~\cite{vaswani2017attention} based mask decoder to produce high-quality segmentation masks. 
Please see Appendix~\ref{sam_pre} for details.

As discussed in Section~\ref{introduction}, multiple works~\cite{mazurowski2023segment,he2023computer} have shown SAM demonstrates reduced effectiveness in downstream tasks when there is  a noticeable difference between the data it was pretrained on and the task-specific data. 
For example, \citet{he2023computer} studied the performance of SAM on 12 medical segmentation datasets, showing SAM performs significantly worse than five SOTA algorithms on some datasets. The findings underscore the necessity of finetuning SAM when it is applied to domains other than natural images. 
To address this, MedSAM~\cite{ma2023segment}, Med-SA~\cite{wu2023medical}, and SAMed~\cite{zhang2023customized} finetune SAM on medical images, outperforming several SOTA methods. 
To enable prompting SAM using texts, LISA~\cite{lai2023lisa} integrates a multi-modal large language model, LLaVA~\cite{liu2023llava}, with SAM. 
In addition, \citet{yang2023track} and \citet{cheng2023tracking} further extend SAM to video segmentation and object 
tracking in videos,  demonstrating SAM's versatility across a range of visual tasks


\subsection{Model Finetuning}
Finetuning techniques have demonstrated state-of-the-art effectiveness in adapting large-scale foundation models for specialized downstream tasks \cite{radford2018improving,devlin2018bert}. However, as foundation models grow in size, finetuning all parameters becomes increasingly expensive \cite{brown2020language,touvron2023llama,kirillov2023segment} in computation.  Therefore, various parameter-efficient finetuning methods have been proposed to alleviate this issue. For instance, Adapter \cite{houlsby2019parameter} injects trainable adapter layers into pretrained Transformers~\cite{vaswani2017attention}. Low-Rank Adaptation (LoRA)~\cite{hu2021lora}  adds learnable low-rank matrices to the pretrained weight matrices of foundation models and only optimizes the low-rank matrices in the finetuning stage with the pretrained parameters frozen. \citet{zhang2023adaptive} proposes to adaptively allocate budgets for updating different LoRA layers based on their importance scores when adapting to a specific downstream task.
Prompt-tuning~\cite{prompt2021lester} proposes optimizable `soft prompts' for a specific downstream task and only optimizes the trainable prompts during finetuning. P-tuning~\cite{ptuning2023} finetunes the pretrained foundation model by training a neural network to generate prompt embeddings while keeping the pretrained parameters frozen. IA3~\cite{2022ia3}  proposes to multiply the output of activation layers in the pretrained foundation models with trainable vectors and optimize these vectors in the finetuning stage.  Contrary to these methods that emphasize parameter efficiency, our approach is centered on mitigating overfitting. It is designed to complement these methods rather than replace them.

\subsection{Bi-Level Optimization}
Bi-level optimization (BLO) refers to a class of optimization problems in which one optimization problem (lower level) is nested within another optimization problem (upper level)~\cite{sinha2017review}. 
Various tasks can be formualted in a BLO framework, including meta-learning~\cite{maml,killamsetty2022nested,qin2023bi}, neural architecture search~\cite{liu2018darts,hosseini2022image}, and data reweighting~\cite{metaweight,garg2022learning,hosseini2023fair}.
In these tasks, model weights are usually learnt during lower-level optimization on the training split of the dataset, while meta-variables like hyper-parameters or architectures are learnt in the upper-level optimization on a separate validation split to alleviate the issue of overfitting.

Numerous gradient-based optimization algorithms and software have been developed for solving BLO problems.
For example, \citet{liu2018darts} develop a finite difference approximation method to efficiently compute the gradients with regard to the upper level variables in BLO problems. \citet{maml} propose to compute the gradient updates of meta variables directly with iterative differentiation \cite{Grazzi2020OnTI}. \citet{choe2022betty} develop a software for users to easily and efficiently compute gradients within BLO problems with different approximation methods. 





\section{Method} \label{method}


\subsection{Overview of BLO-SAM}
Our approach, BLO-SAM, finetunes SAM for downstream semantic segmentation tasks using bi-level optimization (BLO). 
In the pretrained SAM model, certain parameters undergo finetuning, whereas others remain static throughout the finetuning process, as shown in Fig.~\ref{fig:overview}.  Furthermore, to eliminate the need for manual prompts, we learn a prompt embedding vector.  
To combat overfitting, we split the training set into two halves (denoted by $D_1$ and $D_2$), which are used for learning SAM's finetunable parameters and the prompt embedding, respectively. 
In the lower-level of our  BLO framework, SAM's finetunable parameters  (denoted by $W$) are optimized on the sub-dataset  $D_1$, with the prompt embedding (denoted by $A$) tentatively fixed. The optimal solution $W^*(A)$, dependent on $A$, is then passed to the upper level. In the upper-level optimization, $W^*(A)$ is validated on the sub-dataset $D_2$. The validation loss, which is a function of  the prompt embedding $A$, indicates  the generalization performance of the finetuned SAM. We optimize $A$  to minimize this validation loss for reducing the risk of overfitting and improving generalization. 
The two levels of problems share the same form of loss function designed for semantic segmentation. 
The two levels are optimized iteratively until convergence, as shown in Algorithm~\ref{algo}. 

\begin{figure}[t]
    \centering
\includegraphics[width=.9\linewidth]{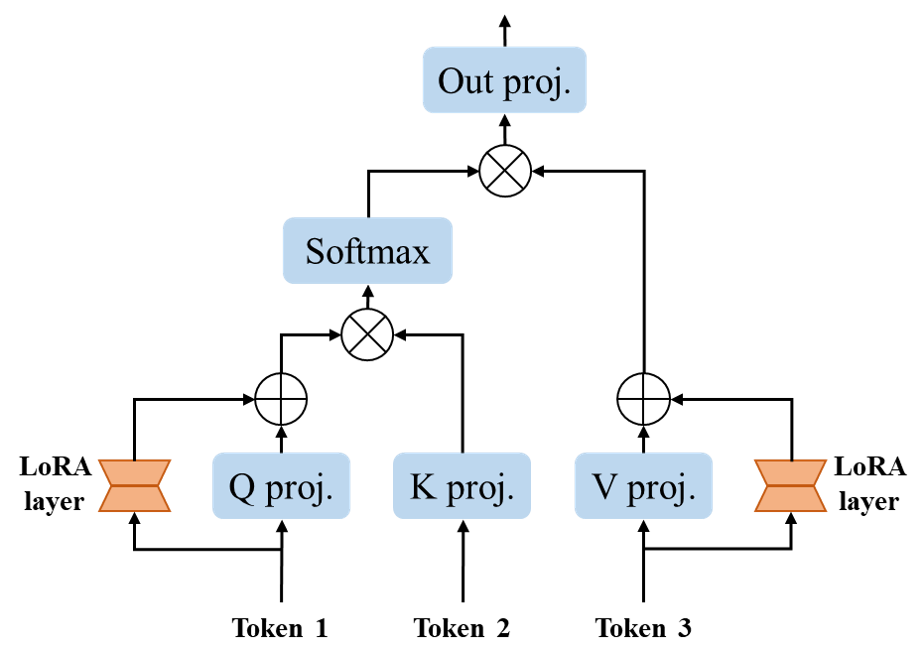}
\vspace{-0.5cm}
\caption{Finetune SAM with LoRA. We add LoRA layers to query and value projection layers within the attention block of SAM's mask decoder. `Q proj.', `K proj.', `V proj.', and `Out proj.' denote the projection layers for the query, key, value, and output.}
\vspace{-0.5cm}
\label{fig:lora-attn}
\end{figure}

\vspace{-0.2cm}
\paragraph{Finetuning SAM with LoRA.} 
We employ Low-Rank Adaptation (LoRA)~\cite{hu2021lora} for parameter-efficient finetuning of SAM. LoRA introduces an additional learnable matrix (known as a LoRA layer), which is of lower rank, as an update to the existing pretrained weight matrix. 
During finetuning, the focus is on optimizing this lower-rank matrix, while keeping the pretrained matrix static. Notably, the LoRA layer comprises considerably fewer parameters than the original matrix, enhancing the efficiency of the finetuning process.  
As shown in Fig.~\ref{fig:lora-attn}, a LoRA layer is added to
each query and value projection layer of all attention blocks in SAM's mask decoder, including self-attention, image-to-token attention, and token-to-image attention. 
Each LoRA layer consists of two sequential linear layers where the first one projects the input token to a low-dimensional space and the second one projects from the low-dimensional space back to the original feature space so that the output of LoRA can be added to the output of the frozen Transformer layer (Fig.~\ref{fig:lora-attn}). 
For different attention blocks, the input tokens differ. In the self-attention block, the prompt embedding serves as all three input tokens. For image-to-token attention blocks, the prompt embedding is used for both key and value projections, while the image embedding is used for the query projection. Conversely, in the token-to-image attention blocks, the image embedding is employed for key and value projections, and the prompt embedding is utilized for the query projection. 
In the end, the learnable parameters of SAM encompass those in the LoRA layers, the transposed convolutions, and the multi-layer perceptron (MLP) head located in the mask decoder, with all other model parameters frozen, as shown in Fig.~\ref{fig:overview}(a).




\subsection{Bi-Level Finetuning Framework}

\paragraph{Lower-Level Optimization Problem.} 
In the lower level, we tentatively fix the prompt embedding $A$ and optimize SAM's finetunable parameters $W$ on the first sub-dataset $D_1$ by minimizing the following loss:
\begin{align} \label{eq: loss}
    \mathcal{L} = (1-\lambda) \mathcal{L}_{CE}(W, A; D_1) + \lambda \mathcal{L}_{Dice}(W, A; D_1),
\end{align}
where $\mathcal{L}_{CE}$ and $\mathcal{L}_{Dice}$ represent cross-entropy loss and dice loss, respectively, between predicted masks and ground-truth masks, and $\lambda$ is a tradeoff parameter. The $(W, A; D_1)$ arguments indicate the loss depends on the model parameters and prompt embedding while computed on dataset $D_1$.
The lower level aims to solve the following problem:
\begin{align} \label{eq:low}
    W^*(A) = \operatorname{\arg\min}_W \mathcal{L}(W, A; D_1).
\end{align}
$W^*(A)$ implies the optimal solution $W^*$ depends on $A$ as $W^*$ depends on the loss function which depends on $A$. 

\vspace{-0.2cm}
\paragraph{Upper-Level Optimization Problem.} In the upper level, we validate the finetuned model $W^*(A)$ on the second sub-dataset $D_2$. The learnable prompt embedding is updated by minimizing the validation loss:
\begin{align} \label{eq:upper}
    \operatorname{\min}_A \mathcal{L}(W^*(A), A; D_2),  
\end{align}
which is the same loss function as Eq.~(\ref{eq: loss}) but with $D_1$ replaced by $D_2$ and $W$ replaced by $W^*(A)$. The key in Eq.~\ref{eq:upper} is that the loss depends on $W^*(A)$ which also depends on $A$. Thus, we need to ``unroll'' $W^*(A)$ to correctly compute the gradient w.r.t.~$A$, as detailed in Appendix~\ref{second-order}.

\vspace{-0.2cm}
\paragraph{Bi-Level Optimization Framework.} Integrating the aforementioned two optimization problems, we have a bi-level optimization framework:
\begin{align}\label{eq:overall}
\min _A &\quad \mathcal{L}(W^*(A), A; D_2) \nonumber \\
s.t. &\quad W^*(A) = \operatorname{\arg\min}_W \mathcal{L}(W, A; D_1)
\end{align}
In this framework, the two optimization problems are interdependent. The output of the lower level, $W^*(A)$, serves as the input for the upper level. The optimization variable $A$ in the upper level is used in the lower-level loss function. 

\subsection{Optimization Algorithm}
We employ a gradient-based optimization algorithm to solve the problem in Eq.~(\ref{eq:overall}). Drawing inspiration from~\citet{liu2018darts}, we perform a one-step gradient descent for Eq.~(\ref{eq:low}) to approximate the optimal solution $W^*(A)$. 
Subsequently, we substitute the approximation of $W^*(A)$ into the upper-level optimization problem. $A$ is updated by minimizing the approximated upper-level loss function 
via gradient descent.  
These steps constitute one global optimization step. We iteratively perform global steps between the lower level and upper level until convergence, as shown in Algorithm~\ref{algo}. Detailed derivations are provided in Appendix~\ref{second-order}.
\begin{algorithm}[t]
    \caption{Optimization Algorithm for BLO-SAM}
    \begin{algorithmic}[1]
    \label{algo}
        \STATE \textbf{Input:} Sub-datasets $D_1$, $D_2$. Learning rates $\eta_1$, $\eta_2$. Parameter initialization $W^{(0)}$, $A^{(0)}$.
        \FOR{$t = 1,2,3,\cdots$}
            \STATE Sample a batch from $D_1$. Update $W^{(t)}$ via Eq.~(\ref{eq:low}).
            \STATE Sample a batch from $D_2$. Update $A^{(t)}$ via Eq.~(\ref{eq:upper}).
        \ENDFOR
    \end{algorithmic}
\end{algorithm}

\begin{table*}[t]
\centering
\vspace{-0.2cm}
\caption{Average Dice score (\%) on the CelebAMask dataset, with different numbers of training examples. Standard deviations are in Appendix~\ref{detailed_res}. 
The overall performance is calculated by averaging the results for different facial components. The last column indicates whether manual prompts are needed.}
\vskip 0.05in
\label{res:face}
\setlength{\tabcolsep}{1.5mm}{
\begin{tabular}{ccccccccccccccc}
\toprule
\multirow{2}{*}{Method} & \multicolumn{6}{c}{Training with 4 labeled examples} & &\multicolumn{6}{c}{Training with 8 labeled examples} & \multirow{2}{*}{\makecell{Manual\\Prompts}}\\ \cline{2-7}\cline{9-14}
& Overall & Brow & Eye & Hair & Nose & Mouth & & Overall & Brow & Eye & Hair & Nose & Mouth \\ \hline\hline
DeepLab & 49.5 & 33.2 & 55.9 & 55.2 & 55.8 & 47.5 & & 54.9 & 37.3 & 59.7 & 58.0 & 64.0 & 55.7 & \XSolidBrush \\
SwinUnet & 35.2 & 21.7 & 21.7 & 46.7 & 44.4 & 41.6 & & 42.4 & 28.8 & 33.1 & 52.7 & 47.6 & 49.8 & \XSolidBrush \\ \hline
HSNet & 49.9 & 30.1 & 41.9 & 58.9 & 59.6 & 59.0 & & 60.1 & 43.6 & 58.1 & 76.6 & 58.2 & 64.2 & \XSolidBrush \\
SSP & 56.9 & \textbf{40.3} & 33.6 & 73.2 & 71.6 & 65.6 & & 60.0 & 45.4 & 31.2 & 76.6 & 74.0 & 72.7 & \XSolidBrush \\ \hline
SAM & 32.9 & 20.8 & 30.6 & 44.0 & 44.1 & 25.2 && 32.9 & 20.8 & 30.6 & 44.0 & 44.1 & 25.2 & \Checkmark \\
Med-SA & 62.9 & 36.3 & 63.6 & 77.6 & 71.0 & 66.0 & & 67.7 & 42.9 & 65.5 & 82.4 & 74.6 & \textbf{73.1} & \Checkmark \\
SAMed & 58.2 & 28.3 & 55.9 & 78.5 & 65.1 & 63.0 & & 65.0 & 39.5 & 66.0 & 82.4 & 70.5 & 66.4 & \XSolidBrush \\ \hline
\textbf{BLO-SAM} & \textbf{65.9} & 39.4 & \textbf{65.5} & \textbf{82.8} & \textbf{74.3} & \textbf{67.6} & & \textbf{69.9} & \textbf{45.8} & \textbf{71.1} & \textbf{83.6} & \textbf{76.1} & 72.7 & \XSolidBrush \\ \bottomrule                  
\end{tabular}}
\vspace{-0.3cm}
\end{table*}

\begin{table*}[t]
\centering
\caption{The comparison of BLO-SAM with baselines on the car components and human body datasets evaluated by Dice Score (\%) with different numbers of training examples. Standard deviations are in Appendix~\ref{detailed_res}. The overall performance is calculated by averaging the results for different components.}
\vskip 0.05in
\label{res: car_body}
\setlength{\tabcolsep}{1.5mm}{
\begin{tabular}{cccccccccc|cc}
\toprule
\multirow{3}{*}{Method} & \multicolumn{9}{c|}{Car components} & \multicolumn{2}{c}{Human body} \\ \cline{2-12}
& \multicolumn{4}{c}{Training with 2 labeled examples} &  & \multicolumn{4}{c|}{Training with 4 labeled examples} & \multirow{2}{*}{4 examples} & \multirow{2}{*}{8 examples} \\ \cline{2-5}\cline{7-10}
 & Overall & Body & Wheel & Window & & Overall & Body & Wheel & Window &  &  \\ \hline \hline
DeepLab & 46.5 & 58.5 & 55.0 & 26.1 & & 59.8 & 62.9 & 66.6 & 49.8 & 31.8 & 37.0 \\
SwinUnet & 31.4 & 22.2 & 33.4 & 38.7 & & 47.3 & 49.3 & 53.7 & 38.9 & 30.9 & 56.8 \\ \hline
HSNet & 51.0 & 67.1 & 32.4 & 53.4 & & 68.8 & 70.8 & 70.3 & 65.4 & 50.6 & 55.8 \\
SSP & 64.1 & 62.8 & \textbf{76.2} & 53.3 & & 72.2 & 77.7 & 78.2 & 60.6 & 58.9 & 76.5 \\ \hline
SAM & 35.1 & 40.8 & 41.7 & 22.9 && 35.1 & 40.8 & 41.7 & 22.9 & 26.0 & 26.0 \\
Med-SA & 67.3 & 80.8 & 70.9 & 50.3 & & 75.9 & 84.4 & 78.1 & 65.3 & 58.8 & 80.6 \\
SAMed & 60.4 & 78.8 & 65.0 & 37.5 & & 74.0 & 85.3 & 72.5 & 64.2 & 63.8 & 81.5 \\ \hline
\textbf{BLO-SAM} & \textbf{71.1} & \textbf{83.2} & 74.5 & \textbf{55.6} & & \textbf{78.3} & \textbf{86.3} & \textbf{79.9} & \textbf{68.6} & \textbf{76.3} & \textbf{85.1} \\ \bottomrule
\end{tabular}}
\vspace{-0.3cm}
\end{table*}

\begin{table*}[t]
\centering
\caption{The comparison of BLO-SAM with baselines on the teeth, gastrointestinal disease and lung datasets evaluated by Dice Score (\%) with different numbers of training examples. Standard deviations are in Appendix~\ref{detailed_res}. The last three columns show the total number of model parameters, the number of trainable parameters (in millions), and the training time (in GPU hours).}
\vskip 0.05in
\label{res: teeth_kvasir_lung}
\setlength{\tabcolsep}{.5mm}{
\begin{tabular}{ccccccccccccc}
\toprule
\multirow{2}{*}{Method} & \multicolumn{2}{c}{Teeth} && \multicolumn{2}{c}{Kvasir} & & \multicolumn{2}{c}{Lung} 
 && \multirow{2}{*}{\makecell{Total\\Param(M)}} & \multirow{2}{*}{\makecell{Trainable\\Param(M)}}  & \multirow{2}{*}{\makecell{Train Cost\\(GPU hours)}}\\ \cline{2-3}\cline{5-6}\cline{8-9}
 & 4 examples & 8 examples && 4 examples & 8 examples & & 2 examples & 4 examples \\ \hline\hline
DeepLab & 56.8 & 63.9 & & 31.9 & 37.8 & & 61.1 & 76.4 && 41.9 & 41.9 & 0.27 \\
SwinUnet & 28.6 & 50.6 & & 37.8 & 39.0 & & 62.1 & 76.4 && 27.2 & 27.2 & 0.29 \\ \hline
HSNet & 70.2 & 72.2 & & 30.0 & 35.1 & & 83.1 & 84.5 && 28.1 & 2.6 & 0.16 \\ 
SSP & 33.7 & 51.7 & & 27.4 & 27.7 & & 83.2 & 90.1 && 8.7 & 8.3 & 0.22 \\ \hline
SAM & 21.2 & 21.2 && 14.7 & 14.7 && 31.4 & 31.4 && 91.0 & 0 & 0 \\
Med-SA & 69.8 & 75.1 &&33.5 & 59.3 & & 82.9 & 91.7 && 104.4 & 10.7 & 0.62 \\
SAMed & 69.9 & 76.4 && 42.7 & 57.1 & & 84.4 & 91.8 && 91.0 & 1.1 & 0.46 \\ \hline
\textbf{BLO-SAM} & \textbf{73.2} & \textbf{77.3} && \textbf{59.7} & \textbf{61.6} &  & \textbf{87.1} & \textbf{93.7} && 91.0 & 1.1 & 0.57 \\ \bottomrule
\end{tabular}}
\vspace{-0.1cm}
\end{table*}

\section{Experiments}
In this section, we evaluate  BLO-SAM across a diverse set of semantic segmentation tasks from both general domains and medical domains, including human face components segmentation, car components segmentation, human body segmentation,  teeth segmentation, gastrointestinal disease segmentation, and lung segmentation. To be compatible with SAM, each multi-class segmentation task is converted to multiple binary segmentation tasks, one for each class. Our experiments focus on ultra-low data regimes, where the number of training examples is less than ten.

\subsection{Datasets}

For the human facial components segmentation task, we employed the CelebAMask-HQ dataset~\cite{lee2020maskgan}, a collection of high-resolution face images accompanied by segmentation masks of various facial components including brow, eye, hair, nose, and mouth. 
For the car segmentation task, we used the dataset from~\citet{car-seg}, which contains images of cars and their segmentation masks with four semantic components: car body, wheel, light, and windows. We only used the car body, wheel, and windows as the `light' category is missing in many data examples. For human body segmentation, we used the TikTok dances dataset~\cite{body-seg}. For teeth, gastrointestinal disease, and lung segmentation tasks, we utilized the children's dental panoramic radiographs dataset~\cite{zhang2023children}, Kvasir-SEG dataset~\cite{jha2020kvasir}, and JSRT dataset~\cite{shiraishi2000development}, respectively. Every dataset is comprised of a training set and a test set. By randomly sampling a small number of examples from the original training set, we create a new few-shot training set which is then used to train our model and baselines. More details about the datasets can be found in Appendix~\ref{data}.
We repeated each sampling three times at random, and the mean performance over the test set is reported in the main paper, while standard deviations are detailed in Appendix~\ref{detailed_res}. 
In our method,  the sampled new training set is further randomly split into two subsets $D_1$ and $D_2$ with equal size. Baseline methods utilize the entire new training set without any subdivision.
 

\subsection{Experimental Settings}

\paragraph{Baselines and Metrics.} We compared our method with a variety of baselines, including supervised, few-shot, and SAM-based approaches. The supervised methods include  DeepLabV3~\cite{chen2017rethinking}, a widely adopted model in semantic segmentation, and SwinUnet~\cite{cao2022swin}, an UNet-like transformer designed for medical image segmentation. Few-shot learning methods include HSNet~\cite{zhang2022hsnet} which uses cross-semantic attention to bridge the gap between low-level and high-level features, and SSP~\cite{fan2022self} which introduces a  self-support matching strategy to capture consistent underlying characteristics of query objects. SAM-based baselines include vanilla  SAM~\cite{kirillov2023segment}, Med-SA~\cite{wu2023medical} and SAMed~\cite{zhang2023customized} which use Adapter~\cite{houlsby2019parameter} and LoRA~\cite{hu2021lora} for SAM finetuning respectively. Med-SA and SAMed update all learnable parameters on a single training set. For vanilla SAM, we constructed three different types of prompts from each ground-truth mask, including a positive point from the foreground, a negative point from the background, and bounding boxes of target objects. For Med-SA, its prompts are extracted from the ground-truth masks as well. 
During inference, both SAM and Med-SA necessitate user-generated prompts, a requirement that is unfeasible when test images are numerous. Following the evaluation protocols of SAM and Med-SA, these prompts are derived in advance from the ground-truth masks. 
It is important to recognize that such an approach is impractical, as it needs to access the ground-truth masks of test images. Nevertheless, they remain the most accessible baselines we can compare against. All experiments are conducted on an A100 GPU with 80G memory.

We used the Dice score as the evaluation metric.
The Dice score is defined as $\frac{2|A \cap B|}{|A|+|B|}$, where $A$ and  $B$ represent the predicted and ground truth mask respectively. 

\vspace{-0.2cm}
\paragraph{Hyper-parameters.} 
In our method, the trade-off parameter $\lambda$ in Eq.~(\ref{eq: loss}) was set to $0.8$ following the setting  in~\citet{zhang2023customized}.  
LoRA layers and other non-frozen model components were optimized in the lower level using the AdamW optimizer~\cite{loshchilov2017decoupled}. We set the initial learning rate to $5e$-3, betas to $(0.9, 0.999)$, and weight decay to $0.1$. The learning rate in  the $i$-th iteration followed the formula~\cite{zhang2023customized}: 
$\text{lr}_i = \text{lr}_0 (1 - i/\text{Iter}_\text{max})^{0.9}$.
Here, $\text{lr}_0$ and $\text{Iter}_\text{max}$ represent the initial learning rate and maximal training iteration, respectively. For updating the prompt embedding in the upper level, we used the same settings as in the lower-level optimization. We set the number of training epochs to $100$ and selected the best checkpoint based on segmentation performance on the $D_2$ sub-dataset. 

\subsection{Results and Analysis}

\paragraph{Human Facial Components Segmentation.} In our evaluation of BLO-SAM for human facial components segmentation on the CelebAMask dataset, we have two settings with different numbers (4 or 8) of training examples. The results presented in Table~\ref{res:face} reveal some noteworthy observations. 

Firstly, BLO-SAM demonstrates its strong capability to finetune SAM for accurate facial component segmentation even with a very small number of training examples. For instance, with just 4 labeled examples (2 for $D_1$ and the other 2 for $D_2$), BLO-SAM achieves an overall Dice score of $65.9\%$, and the score improves to $69.9\%$ when the training dataset size is increased to 8. The effectiveness of BLO-SAM extends to individual facial components, exemplified by the $82.8\%$ Dice score for hair segmentation with only 4 training examples, achieved without any manual input prompts.

Secondly, our BLO-SAM method demonstrates a significant improvement over the standard SAM. For instance,  when finetuned with just 4 labeled examples, 
BLO-SAM dramatically improves the Dice score of SAM from $32.9\%$ to $65.9\%$. 
These results clearly indicate the superiority of our method as an effective finetuning approach for SAM. It adeptly adapts the pretrained SAM to the specific data distributions of downstream tasks, using just a few examples.





Thirdly, BLO-SAM surpasses other SAM-based methods, including Med-SA and SAMed, which also finetune the pretrained SAM for specific semantic segmentation tasks. Unlike Med-SA or SAMed, which update all learnable parameters on a single training set, BLO-SAM employs bi-level optimization to update segmentation model weights and prompt embedding on disjoint subsets of the training data, effectively mitigating the risk of overfitting. While 
Med-SA's performance is close to that of our method, it requires manual prompts from users during testing. This requirement substantially restricts its practicality in real-world applications. In contrast, our method operates entirely autonomously, eliminating the need for manual prompting. 

Finally, compared to general supervised methods such as DeepLab and SwinUnet, BLO-SAM shows a substantial advantage when trained with very few examples. For instance, with 4 labeled examples, DeepLab achieves an overall Dice score of $49.5\%$, whereas BLO-SAM attains a substantially higher Dice score of $65.9\%$. BLO-SAM also outperforms few-shot learning methods including HSNet and SSP, except for brow segmentation with 4 examples. This superiority is attributed to its ability to transfer the capacity of SAM to downstream tasks via finetuning.

\begin{figure}[t]
    \centering
    \includegraphics[width=1\linewidth]{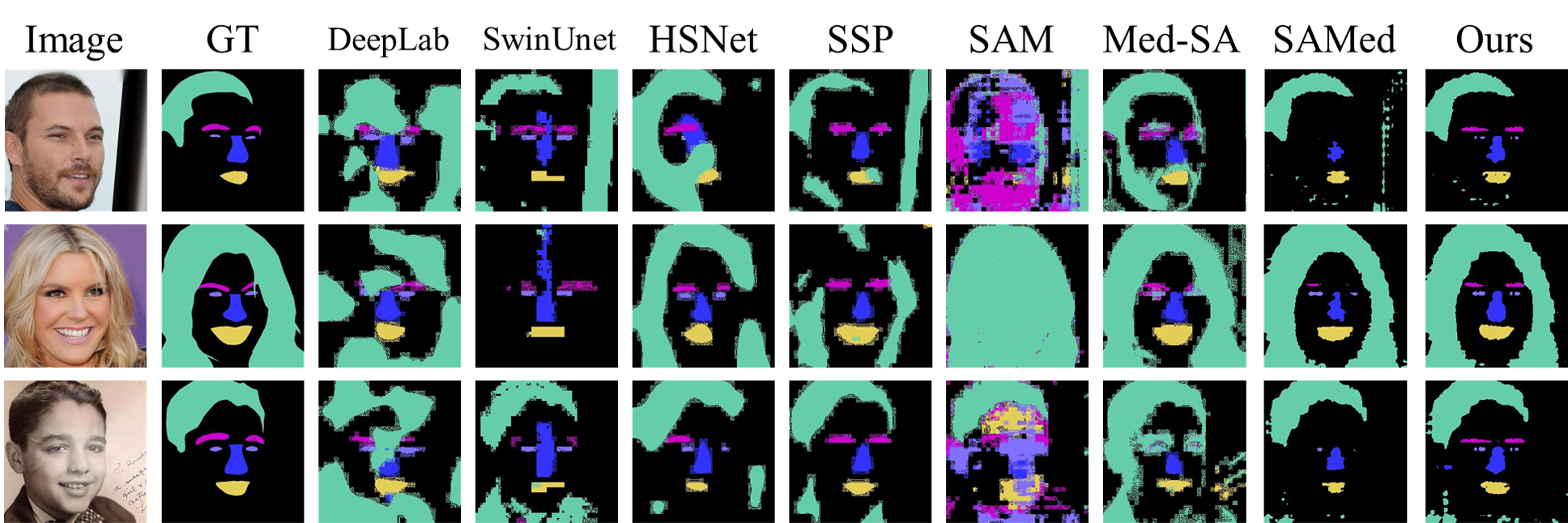}
    \vspace{-0.8cm}
    \caption{Qualitative results on some randomly sampled test images from the CelebAMask dataset. `GT' denotes ground-truth segmentation masks.  
   }
   \vspace{-0.5cm}
    \label{fig: qua_celeb}
\end{figure}

Fig.~\ref{fig: qua_celeb} shows some qualitative comparisons. We can see that BLO-SAM generates more accurate segmentation masks compared to the baselines.

\vspace{-0.2cm}
\paragraph{Car Segmentation and Human Body Segmentation.} 

We further assess BLO-SAM's performance in segmenting car components. Table~\ref{res: car_body} demonstrates that BLO-SAM surpasses all baseline methods in terms of the overall Dice score. For individual components, BLO-SAM excels over the baselines in most categories, with only an exception in `wheel'.  
In the task of human body segmentation, BLO-SAM similarly outperforms the baseline models, as evidenced in Table~\ref{res: car_body}. 
Again, the superiority of our method lies in its strong ability to adapt  SAM to the data distributions of downstream tasks with just a few data examples, while combating overfitting using the BLO framework and eliminating the need for manual prompts.  We further increased the number of  training examples in the body segmentation task. The results are deferred to  Appendix~\ref{train_more}.



\vspace{-0.2cm}
\paragraph{Medical Image Semantic Segmentation.} To further evaluate our method, we perform experiments on several medical image segmentation tasks, where there may be a significant distribution shift from the pretraining data of SAM to the medical datasets. This distribution shift can significantly impair SAM's capability in the medical domain. We perform experiments on teeth, gastrointestinal disease, and lung segmentation tasks and report the results in Table~\ref{res: teeth_kvasir_lung}. For all medical segmentation tasks, BLO-SAM attains the best performance compared with  baselines. The advantage of BLO-SAM over the baselines is more evident in the 4-example case than in the 8-example case, which underscores BLO-SAM's strong ability to combat overfitting and improve generalization in few-shot settings. BLO-SAM's superiority can also be demonstrated by comparing the predicted segmentation masks, as shown in Appendix~\ref{res: qua}. 

\vspace{-0.2cm}
\paragraph{Computational Costs and Parameter Counts.} We measured the training cost for all methods on an A100 GPU. We can see that BLO-SAM has comparable training time (Table~\ref{res: teeth_kvasir_lung}) and inference time (Appendix~\ref{train_test_cost}) to other SAM-based methods.  More detailed analysis can be found in Appendix~\ref{train_test_cost}.  Furthermore, 
as shown in Table~\ref{res: teeth_kvasir_lung}, BLO-SAM has minimal trainable parameters among all trainable methods, underscoring its parameter efficiency.

\subsection{Ablation Studies}

\begin{figure}[t]
    \centering
    \includegraphics[width=1\linewidth]{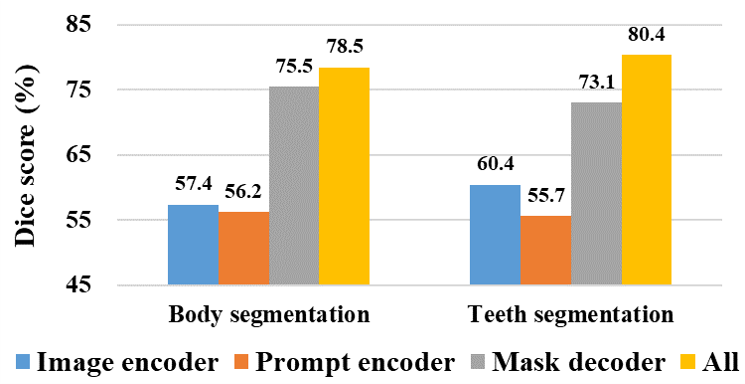}
    \vspace{-0.8cm}
    \caption{Ablation study of different trainable components. ``All" represents finetuning all three components of SAM. 
    }
    \vspace{-0.3cm}
    \label{fig: abla1}
\end{figure}

\paragraph{Ablation of Trainable Components of SAM.} 
SAM comprises three primary components: image encoder, prompt encoder, and mask decoder. We have 4 ablation settings, including 1) finetuning the mask decoder (same as BLO-SAM), 2) the image encoder (via LoRA layers that are integrated into the inner Transformer blocks), 3) the prompt encoder (by updating the convolution layers), and 4) finetuning all the above components. 
This ablation experiment was conducted on body and teeth segmentation datasets, each with 4 labeled examples during finetuning.  The results are  shown in Fig.~\ref{fig: abla1}, where we have several 
key observations. Firstly, the mask decoder stands out as the most pivotal component for finetuning when adapting SAM to specific semantic segmentation tasks.  Secondly, finetuning all three components of SAM yields improved performance, albeit at the expense of increased trainable parameters and training cost. For this reason, we only finetune the mask decoder in our BLO-SAM approach.

\begin{figure}[t]
    \centering
    \includegraphics[width=.9\linewidth]{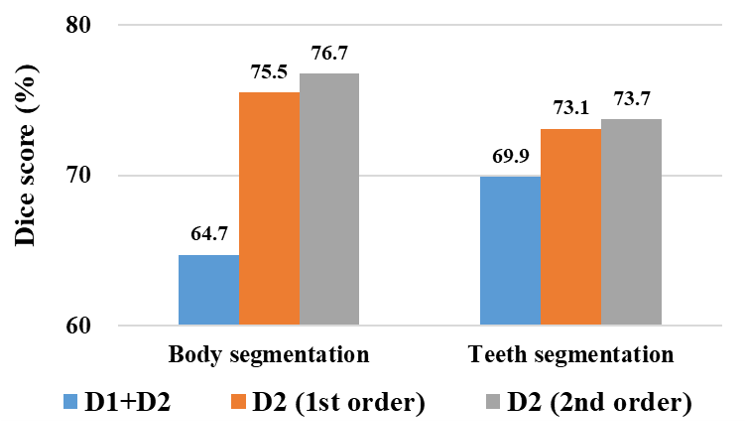}
    \vspace{-0.3cm}
    \caption{Ablation study on approaches  for optimizing the prompt embedding. 
    }
    \vspace{-0.4cm}
    \label{fig: abla2}
\end{figure}

\paragraph{Ablation on Methods for Optimizing Prompt Embedding.}
In this study, we investigate the effectiveness of three approaches for optimizing prompt embedding. The experiments are performed on the body and teeth segmentation tasks with 4 labeled examples. 
The first approach optimizes both the prompt embedding $A$ and model parameters $W$ on the combination of $D_1$ and $D_2$. 
The second and third approaches optimize the prompt embedding on $D_2$ as in BLO-SAM but use  first-order and second-order approximation, respectively, to compute the gradients in the BLO problem (see Appendix~\ref{second-order} for details). In the previously   mentioned experiments, we use  first-order approximation, which is computationally more efficient than the second-order approximation, to reduce  computational cost. 

Analysis of the results presented in Fig.~\ref{fig: abla2} reveals that optimizing parameters $A$ on sub-dataset $D_2$, while concurrently optimizing parameters $W$ on sub-dataset $D_1$, yields superior performance compared to the simultaneous optimization of both parameter sets $A$ and $W$ on the combination of  $D_1$ and $D_2$. This observation underscores the advantage of separately optimizing distinct parameter sets on disjoint sub-datasets, which is demonstrated to be more effective in mitigating the risk of overfitting compared to the approach of optimizing all parameters on a single dataset. 
Moreover, the results show that second-order optimization holds the potential to further improve the final performance of the model, compared with the first-order method. However, it is crucial to acknowledge that these benefits come with computational challenges and necessitate increased computational resources. 

Appendix~\ref{other_abla} presents three additional ablation studies. 

\vspace{-0.1cm}
\section{Conclusion}
\vspace{-0.1cm}

In this paper, we propose BLO-SAM, a new approach for finetuning the Segment Anything Model (SAM) to perform downstream semantic segmentation tasks without requiring manual prompting.  Leveraging a bi-level optimization strategy, we optimize the segmentation model parameters and prompt embedding on two different sub-datasets, mitigating the risk of overfitting and enhancing generalization.  Experiments across diverse tasks with limited labeled training data strongly demonstrate the effectiveness of BLO-SAM. 


\newpage
\section{Impact Statements}
The paper that focuses on finetuning the Segment Anything Model (SAM) represents a significant stride in advancing semantic segmentation in few-shot settings, particularly in the context of both general and medical domains. Ethically, this endeavor raises questions about the responsible use of powerful AI technologies in healthcare, potentially improving diagnostic processes but also necessitating careful consideration of patient privacy and data security. The societal implications are broad, as the improved segmentation capabilities can benefit diverse industries, from autonomous vehicles to medical imaging. However, the responsible deployment of such advancements is crucial to avoid unintended consequences and biases. Striking a balance between innovation and ethical considerations will be pivotal in harnessing the full potential of the research for the betterment of society.


\bibliography{main}

\newpage
\appendix
\onecolumn

\section{Detailed Optimization Algorithm} \label{second-order}
In this section, we offer a detailed description of the optimization algorithm of BLO-SAM. We employ a gradient-based optimization algorithm to tackle the problem outlined in Eq.~\ref{eq:overall}. Drawing inspiration from~\citet{liu2018darts}, we approximately update $W^*(A)$ via one-step gradient descent in the lower level optimization. Then we plug the approximate $W^*(A)$ into the learning process of prompt embedding in the upper level and update $A$ via one-step gradient descent. By using the one-step gradient descent updates for the bi-level optimization framework, we reduce the computational complexity. The detailed derivation of the update is as follows.

\paragraph{Lower-level.} 
For the lower level, we perform a one-step gradient descent for Eq.~(\ref{eq:low}) to approximate the optimal solution $W^*(A)$ on $D_1$. Specifically, at step $t$ with an initial $W^{(t)}$ and a learning rate $\eta_1$, the updated $W^{(t+1)}$ is computed via gradient descent as follows:
\begin{align} \label{eq:update_low}
    W^{(t+1)} = W^{(t)} - \eta_1 \frac{d \mathcal{L}(W^{(t)}, A^{(t)}; D_1)}{d W^{(t)}}
\end{align}

\paragraph{Upper-level.}
Subsequently, we substitute $W^{(t+1)}$ as an approximation $W^*(A)$ into the upper-level optimization problem. Employing a similar one-step gradient descent, we approximate the optimal solution for $A$ by minimizing the loss on $D_2$. At step $t$ with an initial $A^{(t)}$ and a learning rate $\eta_2$, the updated $A^{(t+1)}$ is calculated as follows:
\begin{align} \label{eq:update_up}
    A^{(t+1)} = A^{(t)} - \eta_2 \frac{d \mathcal{L}(W^*(A), A^{(t)}; D_2)}{d A^{(t)}}
\end{align}
Inspired by~\citet{liu2018darts}, we apply the unrolled model for the parameters of prompt embedding, $A$. In such a setting, the gradient w.r.t. $A$ is:
\begin{align}
    \nabla_{A}\mathcal{L}_{D_2}(W^*(A), A) \approx \nabla_{A}\mathcal{L}_{D_2}(W-\xi \nabla_{W}\mathcal{L}_{D_1}(W, A), A)
\end{align}
where $\xi$ is the learning rate of the lower-level optimization problem. Applying the chain rule to the approximate gradient yields:
\begin{align} \label{eq:second}
    &\nabla_{A}\mathcal{L}_{D_2}(W-\xi \nabla_{W}\mathcal{L}_{D_1}(W, A), A) \nonumber \\ 
    = &\nabla_{A}\mathcal{L}_{D_2}(W^*, A) - \xi \nabla^2_{A,W}\mathcal{L}_{D_1}(W, A) \cdot \nabla_{W^*}\mathcal{L}_{D_2}(W^*, A)
\end{align}
for the second part of Eq.~(\ref{eq:second}), we can apply the infinite difference approximation to simplify it to be: 
\begin{align}
    &\nabla_{A}\mathcal{L}_{D_2}(W^*, A) - \xi \nabla^2_{A,W}\mathcal{L}_{D_1}(W, A) \cdot \nabla_{W^*}\mathcal{L}_{D_2}(W^*, A) \nonumber \\
    \approx & \frac{\nabla_{A}\mathcal{L}_{D_1}(W^+, A) - \nabla_{A}\mathcal{L}_{D_1}(W^-, A)}{2\epsilon}
\end{align}
where, $W^\pm=W \pm \epsilon\nabla_{W^*}\mathcal{L}_{D_2}(W^*, A)$, and $\epsilon = 0.01/\| \nabla_{W^*}\mathcal{L}_{D_2}(W^*, A) \|_2$. If we set the $\xi$ in Eq.~(\ref{eq:second}) to be 0, we can get the first-order optimization for prompt embedding, otherwise we get the second-order optimization for the prompt embedding. For our main method, we utilize the first-order optimization method for its low computational cost. In the ablation studies in the main paper, we analyze the effect of using the second-order optimization method.

\section{Preliminaries of Segment Anything Model (SAM)}\label{sam_pre}
\begin{figure}[t]
    \centering
    \includegraphics[width=1\linewidth]{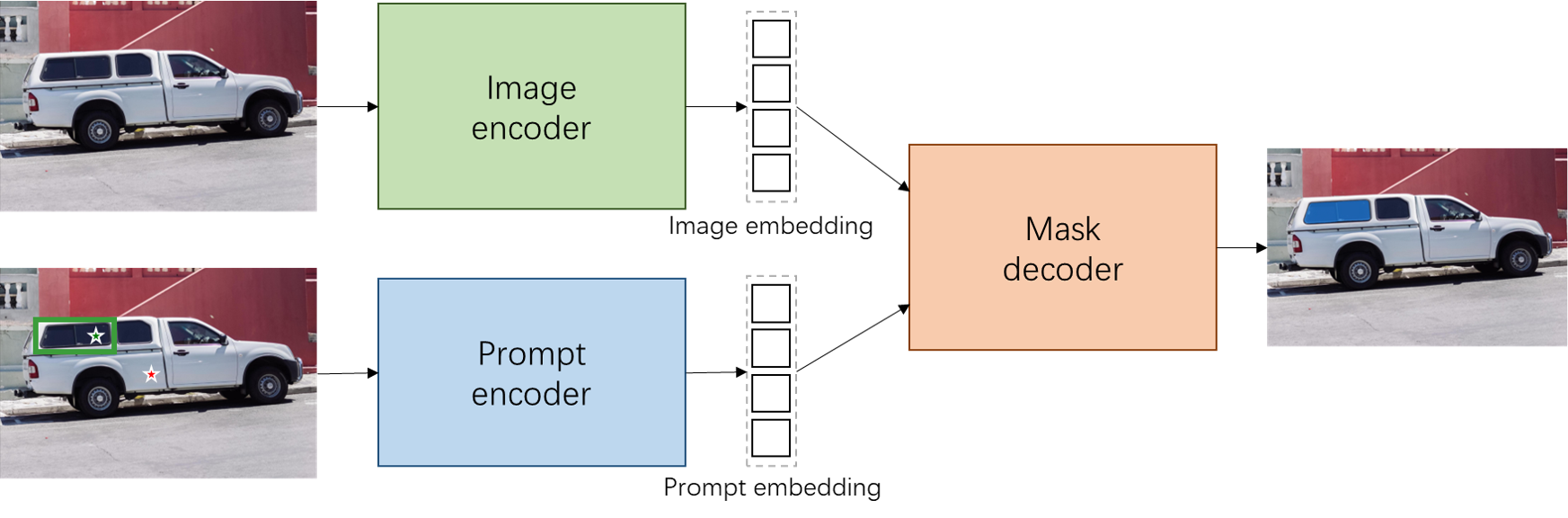}
    \vspace{-0.5cm}
    \caption{SAM overview. The image encoder takes an image as input to get the corresponding image embedding. The prompt encoder takes the input prompts, indicated as green point, red point and the rectangle shown in the image, to generate the prompt embedding. The mask decoder takes the image and prompts embedding as input to output the final masks.}
    \vspace{-0.3cm}
    \label{fig: sam}
\end{figure}

The Segment Anything Model (SAM)~\cite{kirillov2023segment} follows a comprehensive workflow, as shown in Fig.~\ref{fig: sam}, beginning with the encoding of an input image and a prompt that indicates the object to segment. 
SAM comprises three key components: an image encoder that processes the input image and generates an image embedding, which captures the visual features of the input image; a prompt encoder that encodes the provided prompts, which captures the semantics of the object or region to segment; and a lightweight mask decoder that takes both image and prompt features as input and generates a segmentation mask for the object or region described in the prompt. The image encoder is a ViT~\cite{dosovitskiy2020image}, which outputs a sequence of image tokens (vectors).
For the prompts, SAM accommodates various types of prompts, such as points, bounding boxes, and coarse masks. Point or bounding box prompts are represented by positional encodings~\cite{tancik2020fourier}. Mask prompts are encoded using a convolutional neural network. 
If no prompts are manually provided to SAM, it dynamically uses a default embedding as the input prompt. The mask decoder, which is a modification of a Transformer decoder block~\cite{vaswani2017attention}, efficiently maps the image embedding, prompt embeddings, and an output token to a mask. In conclusion, SAM is a powerful and versatile tool for promptable segmentation, capable of handling a wide range of segmentation tasks efficiently and effectively.
There are three model versions of the SAM with three different types of image encoders, scaling from ViT-B, ViT-L and ViT-H. ViT-B represents the encoder with the smallest model size. And, in our experiments, we use the ViT-B for computational efficiency by default.

\begin{table}[t]
\centering
\vspace{-0.1cm}
\caption{Number of test examples in different tasks.}
\label{tab: data}
\setlength{\tabcolsep}{1mm}{
    \begin{tabular}{l|cccccc}
    \hline
        Dataset & CelebAMask & Car & Body  & Teeth & Kvasir & Lung \\ 
    \hline
        Test size & 2000 & 100 & 2000 & 70 & 500 & 147 \\
    \hline
    \end{tabular}
    }
    \vspace{-0.3cm}
\end{table}

\section{Datasets} \label{data}
We evaluate our method on six tasks, including three tasks in the general domain, and three tasks in the medical domain. Correspondingly, we used a total of six datasets in our experiments, each of which was publicly available. The statistics of the test sets are listed in Table~\ref{tab: data}.

For the facial components segmentation task, we use the CelebAMask-HQ dataset~\footnote{https://drive.google.com/open?id=1badu11NqxGf6qM3PTTooQDJvQbejgbTv}, which is a large-scale face image dataset that has 30,000 high-resolution face images selected from the CelebA dataset by following CelebA-HQ. Each image has its segmentation mask of facial attributes corresponding to CelebA. We use the last 2,000 examples as the test set.
For the car components segmentation task, we use the Car segmentation dataset~\footnote{https://www.kaggle.com/datasets/intelecai/car-segmentation}, which contains 211 images of cars and their segmentation masks. The images exhibit a diverse range of car models and environmental contexts, ensuring that our segmentation method is tested across various visual scenarios. We split the last 100 examples as the test set. 
For the human body segmentation task, we use the Human Segmentation Dataset - TikTok Dances~\footnote{https://www.kaggle.com/datasets/tapakah68/segmentation-full-body-tiktok-dancing-dataset}, which includes 2615 images of a segmented dancing people that are extracted from the videos from TikTok. We split the last 2000 examples as the test set, ensuring a robust evaluation of our method across a spectrum of dance scenarios.

For the teeth segmentation task, we use the Children's Dental Panoramic Radiographs Dataset~\footnote{https://www.kaggle.com/datasets/truthisneverlinear/childrens-dental-panoramic-radiographs-dataset/data}, which is the world’s first dataset of children’s dental panoramic radiographs for caries segmentation and dental disease detection by segmenting and detecting annotations is published in this dataset. This dataset has already split its original examples into train and test sets, thus we just follow the original setting.
For the gastrointestinal disease segmentation task, we use the Kvasir dataset~\footnote{https://www.kaggle.com/datasets/abdallahwagih/kvasir-dataset-for-classification-and-segmentation}, which contains 1000 polyp images and their corresponding ground truth. The diverse set of polyp images captures various shapes, sizes, and locations within the gastrointestinal tract, contributing to the robustness of our segmentation algorithm. We split the last 500 examples as the test set.
For lung segmentation, we utilize the JSRT (Japanese Society of Radiological Technology) database~\footnote{http://db.jsrt.or.jp/eng.php}, containing 247 chest radiographs annotated with precise lung masks. This valuable dataset enhances the robustness of our segmentation method, capturing diverse clinical scenarios. The last 147 examples serve as our test set, ensuring a comprehensive evaluation of our algorithm's performance.

\section{Training and Inference Cost} \label{train_test_cost}

In Table~\ref{tab:cost}, we show the training and inference cost comparison for all methods on the teeth segmentation task that are trained with 4 labeled examples. All the experiments are conducted on an A100 GPU with 80G memories. We use the GPU hours/seconds to measure the time cost for training and inference, respectively. From the results, we can see that BLO-SAM can achieve comparable performance in the training time, and BLO-SAM is among the smallest inference cost groups upon the inference time. 
Specifically, from the comparison of SAMed and BLO-SAM, We can see that the application of the bi-level optimization strategy will increase the training time slightly, but it will not influence the inference time at all. And, compared with other SAM-based methods (e.g. vanilla SAM and Med-SA), BLO-SAM also shows a superior inference time, because it doesn't need to input the prompts manually. Finally, compared with other baselines, including the supervised and few-shot methods, BLO-SAM shows slightly inferior performance in both training and inference costs, but BLO-SAM also shows a much better semantic segmentation performance.

\begin{table}[ht]
    \centering
    \caption{The training and inference time cost comparisons for the teeth segmentation task with 4 examples.}
     \label{tab:cost}
    \begin{tabular}{c|ccccccccc}
    \toprule
         Method & DeepLab & SwinUnet & HSNet & SSP & SAM & Med-SA & SAMed & BLO-SAM \\ \hline
         Training (GPU hours) & 0.27 & 0.29 & 0.16 & 0.22 & 0 & 0.62 & 0.46 & 0.57\\
         Inference (GPU seconds) & 9.5 & 9.6 & 14.7 & 37.9 & 11.2 & 32.9 & 10.4 & 10.4\\ \bottomrule
    \end{tabular}
\end{table}

\section{Detailed Results} \label{detailed_res}
We present a comprehensive analysis of all results, including mean and standard values, in Table~\ref{res:face_detailed}, Table~\ref{res: car_body_detailed}, and Table~\ref{res: teeth_kvasir_lung_detailed}. A notable trend emerges as we observe the standard values: consistently, BLO-SAM outperforms the baselines by exhibiting smaller standard deviations across the majority of experiments. This trend indicates that BLO-SAM demonstrates enhanced stability compared to the baselines. The smaller standard deviations underscore the method's robustness and reliability, showcasing its ability to consistently yield precise results across diverse segmentation tasks.

\begin{table*}[t]
\centering
\caption{The comparison of BLO-SAM with baselines on the CelebAMask dataset evaluated by Dice Score (\%), with different numbers of training examples. The overall performance is calculated by averaging the results for different facial components.}
\vskip 0.1in
\label{res:face_detailed}
\begin{tabular}{ccccccc}
\toprule
\multirow{2}{*}{Method} & \multicolumn{6}{c}{Training with 4 labeled examples} \\ \cline{2-7}
& Overall & Brow & Eye & Hair & Nose & Mouth \\ \hline\hline
DeepLab & 49.5$\pm$3.5 & 33.2$\pm$0.7 & 55.9$\pm$2.9 & 55.2$\pm$5.7 & 55.8$\pm$3.1 & 47.5$\pm$5.3 \\
SwinUnet & 35.2$\pm$1.8 & 21.7$\pm$1.8 & 21.7$\pm$2.5 & 46.7$\pm$1.6 & 44.4$\pm$2.0 & 41.6$\pm$0.9 \\ \hline
HSNet & 49.9$\pm$1.9 & 30.1$\pm$2.9 & 41.9$\pm$1.5 & 58.9$\pm$2.2 & 59.6$\pm$1.6 & 59.0$\pm$1.1 \\
SSP & 56.9$\pm$0.6 & \textbf{40.3$\pm$0.6} & 33.6$\pm$1.0 & 73.2$\pm$0.3 & 71.6$\pm$0.5 & 65.6$\pm$0.8 \\ \hline
SAM & 32.9$\pm$2.3 & 20.8$\pm$2.4 & 30.6$\pm$1.9 & 44.0$\pm$1.7 & 44.1$\pm$1.0 & 25.2$\pm$4.7 \\
Med-SA & 62.9$\pm$0.9 & 36.3$\pm$1.2 & 63.6$\pm$0.6 & 77.6$\pm$0.9 & 71.0$\pm$0.9 & 66.0$\pm$1.1 \\
SAMed & 58.2$\pm$1.5 & 28.3$\pm$1.4 & 55.9$\pm$2.7 & 78.5$\pm$1.3 & 65.1$\pm$1.8 & 63.0$\pm$0.5 \\ \hline
\textbf{BLO-SAM} & \textbf{65.9$\pm$0.6} & 39.4$\pm$0.9 & \textbf{65.5$\pm$0.4} & \textbf{82.8$\pm$0.8} & \textbf{74.3$\pm$0.8} & \textbf{67.6$\pm$0.1}  \\ \bottomrule  \bottomrule 
\multirow{2}{*}{Method} & \multicolumn{6}{c}{Training with 8 labeled examples} \\ \cline{2-7}
& Overall & Brow & Eye & Hair & Nose & Mouth \\ \hline\hline
DeepLab & 54.9$\pm$1.4 & 37.3$\pm$0.7 & 59.7$\pm$1.9 & 58.0$\pm$2.3 & 64.0$\pm$1.1 & 55.7$\pm$0.8 \\
SwinUnet & 42.4$\pm$1.4 & 28.8$\pm$1.5 & 33.1$\pm$1.7 & 52.7$\pm$1.6 & 47.6$\pm$0.9 & 49.8$\pm$1.3 \\ \hline
HSNet & 60.1$\pm$1.9 & 43.6$\pm$2.3 & 58.1$\pm$0.7 & 76.6$\pm$2.9 & 58.2$\pm$2.9 & 64.2$\pm$0.8 \\
SSP & 60.0$\pm$0.9 & 45.4$\pm$0.7 & 31.2$\pm$0.2 & 76.6$\pm$1.4 & 74.0$\pm$1.2 & 72.7$\pm$0.9 \\ \hline
SAM & 32.9$\pm$2.1 & 20.8$\pm$2.4 & 30.6$\pm$0.9 & 44.0$\pm$1.7 & 44.1$\pm$1.0 & 25.2$\pm$4.7 \\
Med-SA & 67.7$\pm$0.9 & 42.9$\pm$0.7 & 65.5$\pm$1.5 & 82.4$\pm$0.6 & 74.6$\pm$0.8 & 73.1$\pm$0.7 \\
SAMed & 65.0$\pm$1.0 & 39.5$\pm$1.4 & 66.0$\pm$0.6 & 82.4$\pm$0.6 & 70.5$\pm$1.2 & 66.4$\pm$1.2 \\ \hline
\textbf{BLO-SAM} & \textbf{69.9$\pm$0.7} & \textbf{45.8$\pm$0.6} & \textbf{71.1$\pm$1.2} & \textbf{83.6$\pm$0.5} & \textbf{76.1$\pm$1.3} & \textbf{72.7$\pm$0.1}  \\ \bottomrule
\end{tabular}
\end{table*}

\begin{table*}[t]
\centering
\caption{The comparison of BLO-SAM with baselines on the car components (left of the vertical bar) and human body (right of the vertical bar) datasets evaluated by Dice Score (\%) with different numbers of training examples. The overall performance is calculated by averaging the results for different components.}
\vskip 0.1in
\label{res: car_body_detailed}
\setlength{\tabcolsep}{.9mm}{
\begin{tabular}{cccccccccc|cc}
\toprule
\multirow{3}{*}{Method} & \multicolumn{9}{c|}{Car components} & \multicolumn{2}{c}{Human body} \\ \cline{2-12}
& \multicolumn{4}{c}{Training with 2 labeled examples} &  & \multicolumn{4}{c|}{Training with 4 labeled examples} & \multirow{2}{*}{4 examples} & \multirow{2}{*}{8 examples} \\ \cline{2-5}\cline{7-10}
 & Overall & Body & Wheel & Window & & Overall & Body & Wheel & Window &  &  \\ \hline \hline
DeepLab & 46.5$\pm$4.6 & 58.5$\pm$5.6 & 55.0$\pm$3.2 & 26.1$\pm$5.0 & & 59.8$\pm$1.5 & 62.9$\pm$1.4 & 66.6$\pm$0.8 & 49.8$\pm$2.2 & 31.8$\pm$2.3 & 37.0$\pm$0.8 \\
SwinUnet & 31.4$\pm$6.8 & 22.2$\pm$6.5 & 33.4$\pm$5.2 & 38.7$\pm$8.7 & & 47.3$\pm$3.1 & 49.3$\pm$5.6 & 53.7$\pm$1.7 & 38.9$\pm$2.0 & 30.9$\pm$3.6 & 56.8$\pm$0.5 \\ \hline
HSNet & 51.0$\pm$1.5 & 67.1$\pm$2.2 & 32.4$\pm$1.0 & 53.4$\pm$1.2 & & 68.8$\pm$0.9 & 70.8$\pm$0.8 & 70.3$\pm$0.8 & 65.4$\pm$1.0 & 50.6$\pm$3.1 & 55.8$\pm$2.4 \\
SSP & 64.1$\pm$1.0 & 62.8$\pm$1.1 & \textbf{76.2$\pm$1.1} & 53.3$\pm$0.9 & & 72.2$\pm$0.9 & 77.7$\pm$0.3 & 78.2$\pm$0.5 & 60.6$\pm$1.8 & 58.9$\pm$0.1 & 76.5$\pm$0.1 \\ \hline
SAM & 35.1$\pm$1.6 & 40.8$\pm$2.5 & 41.7$\pm$1.1 & 22.9$\pm$1.2 && 35.1$\pm$1.6 & 40.8$\pm$2.5 & 41.7$\pm$1.1 & 22.9$\pm$1.2 & 26.0$\pm$1.3 & 26.0$\pm$1.3 \\
Med-SA & 67.3$\pm$2.1 & 80.8$\pm$1.5 & 70.9$\pm$0.5 & 50.3$\pm$4.2 & & 75.9$\pm$1.3 & 84.4$\pm$1.0 & 78.1$\pm$0.3 & 65.3$\pm$2.5 & 58.8$\pm$0.7 & 80.6$\pm$0.4 \\
SAMed & 60.4$\pm$1.5 & 78.8$\pm$1.4 & 65.0$\pm$1.6 & 37.5$\pm$1.6 & & 74.0$\pm$1.4 & 85.3$\pm$0.6 & 72.5$\pm$0.6 & 64.2$\pm$3.0 & 63.8$\pm$1.3 & 81.5$\pm$1.2 \\ \hline
\textbf{BLO-SAM} & \textbf{71.1$\pm$0.7} & \textbf{83.2$\pm$0.6} & 74.5$\pm$0.5 & \textbf{55.6$\pm$1.0} & & \textbf{78.3$\pm$0.5} & \textbf{86.3$\pm$0.1} & \textbf{79.9$\pm$0.9} & \textbf{68.6$\pm$0.4} & \textbf{76.3$\pm$0.8} & \textbf{85.1$\pm$0.5} \\ \bottomrule
\end{tabular}}
\end{table*}

\begin{table*}[t]
\centering
\caption{The comparison of BLO-SAM with baselines on the teeth, gastrointestinal disease and lung datasets evaluated by Dice Score (\%) with different numbers of training examples.}
\vskip 0.1in
\label{res: teeth_kvasir_lung_detailed}
\begin{tabular}{ccccccccc}
\toprule
\multirow{2}{*}{Method} & \multicolumn{2}{c}{Teeth} && \multicolumn{2}{c}{Kvasir} & & \multicolumn{2}{c}{Lung} 
 \\ \cline{2-3}\cline{5-6}\cline{8-9}
 & 4 examples & 8 examples && 4 examples & 8 examples & & 2 examples & 4 examples \\ \hline\hline
DeepLab & 56.8$\pm$0.3 & 63.9$\pm$1.0 & & 31.9$\pm$0.7 & 37.8$\pm$0.3 & & 61.1$\pm$0.3 & 76.4$\pm$0.5 \\
SwinUnet & 28.6$\pm$1.6 & 50.6$\pm$1.1 & & 37.8$\pm$0.3 & 39.0$\pm$0.8 & & 62.1$\pm$0.1 & 76.4$\pm$0.3 \\ \hline
HSNet & 70.2$\pm$1.3 & 72.2$\pm$0.4 & & 30.0$\pm$0.7 & 35.1$\pm$0.8 & & 83.1$\pm$0.5 & 84.5$\pm$0.4 \\ 
SSP & 33.7$\pm$0.5 & 51.7$\pm$1.2 & & 27.4$\pm$0.3 & 27.7$\pm$0.4 & & 83.2$\pm$0.3 & 90.1$\pm$0.6 \\ \hline
SAM & 21.2$\pm$1.6 & 21.2$\pm$1.6 && 14.7$\pm$3.8 & 14.7$\pm$3.8 && 31.4$\pm$0.1 & 31.4$\pm$0.1 \\
Med-SA & 69.7$\pm$0.3 & 76.2$\pm$0.4 && 33.5$\pm$3.5 & 59.3$\pm$0.4 & & 82.9$\pm$0.5 & 91.7$\pm$0.4 \\
SAMed & 69.8$\pm$0.5 & 75.1$\pm$2.2 && 42.7$\pm$0.5 & 57.1$\pm$0.7 & & 84.4$\pm$0.5 & 91.8$\pm$0.4 \\ \hline
\textbf{BLO-SAM} & \textbf{73.2$\pm$0.7} & \textbf{77.3$\pm$0.2} && \textbf{59.7$\pm$0.2} & \textbf{61.6$\pm$0.3} &  & \textbf{87.1$\pm$0.3} & \textbf{93.7$\pm$0.1} \\ \bottomrule
\end{tabular}
\end{table*}

\section{Other Ablations} \label{other_abla}
In this section, we show some results for other ablation studies, without other declarations, all the ablation studies are conducted on the body and teeth segmentation tasks with 4 labeled examples in default.

\paragraph{Sensitivity Analysis of the  Trade-off Parameter $\lambda$.} \label{abla: lambda}
In this experiment, we investigate how different settings of the trade-off parameter ($\lambda$) in Eq.~\ref{eq: loss} affect the model's final performance.
Analyzing the results presented in Fig.~\ref{fig: abla3}, it is evident that setting $\lambda$ to a value in the middle ground yields the optimal performance for both tasks. It is important to balance the two losses as the cross-entropy loss primarily concentrates on the classification for each pixel independently, making it susceptible to overfitting in the presence of class imbalance, where images may consist of a substantial number of background pixels. In contrast, the Dice loss mitigates this issue by prioritizing spatial overlap. 
These results show the importance of striking a balance between pixel-level classification accuracy and spatial overlap. The default value of 0.8 suggested by~\cite{zhang2023customized} is a good choice.

\paragraph{Ablation of the Setting the Rank of the LoRA Layers.}
Analyzing the results in Fig.~\ref{fig: abla4}, it becomes apparent that, for body and teeth segmentation tasks, setting the number of ranks to 4 yields the best generalization on test examples. The choice of the number of ranks is primarily determined by the complexity of specific tasks; increasing the rank excessively can lead to a more complex model that overfits a small training set and fails to generalize to unseen test examples.

\paragraph{Comparison with Full Finetuning. }In this experiment, we compare our method with full finetuning, denoting as FT-SAM, in which we set all the parameters of vanilla SAM to be trainable and tuned them without any modification. As shown in Table~\ref{fig: abla5}, FT-SAM suffers severe overfitting on the body segmentation task, in which FT-SAM only achieves the dice score of $47.5\%$, while BLO-SAM achieves the dice score of $75.5\%$. This is mainly because, with very limited training examples, full finetuning is easy to overfit to the training examples, causing a very poor generalization on the test examples. And, for the teeth segmentation task, BLO-SAM can also achieve comparable performance with the FT-SAM on the test set.

\section{Train with More Examples} \label{train_more}
\begin{figure}[t]
\centering
\begin{minipage}{.45\linewidth}
\centering
\includegraphics[width=1.0\textwidth]{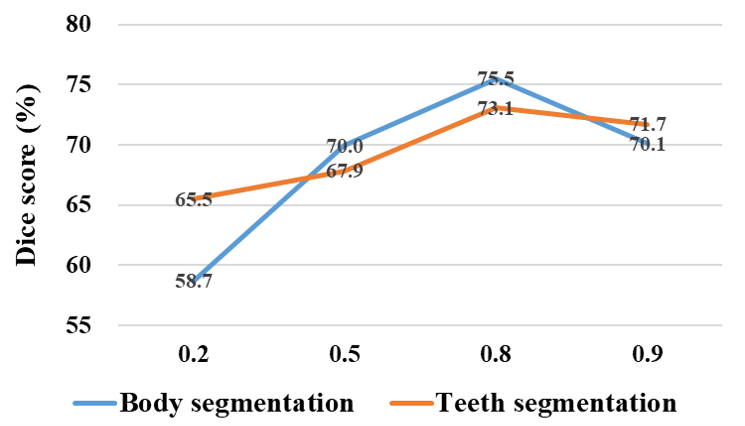}
\vspace{-0.8cm}
\caption{Ablation study of optimization methods for prompt embedding.}
\label{fig: abla3}
\end{minipage}\quad
\begin{minipage}{.45\linewidth}
\centering
\includegraphics[width=1.0\textwidth]{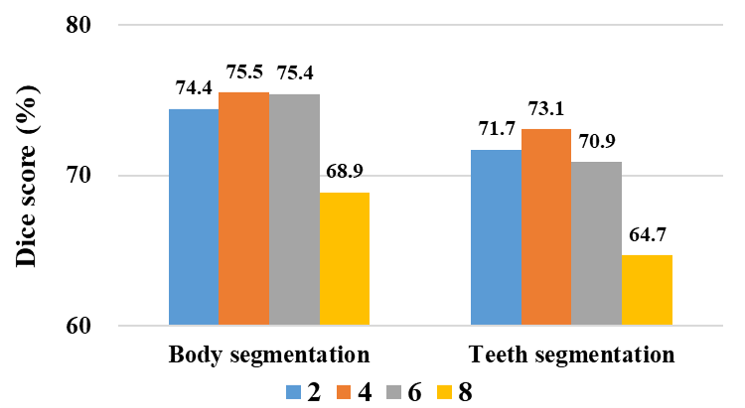}
\vspace{-0.8cm}
\caption{Ablation study of the setting of the number of ranks for added LoRA layers. }
\label{fig: abla4}
\end{minipage}
\vspace{0.3cm}
\begin{minipage}{.45\linewidth}
\centering
\includegraphics[width=.9\textwidth]{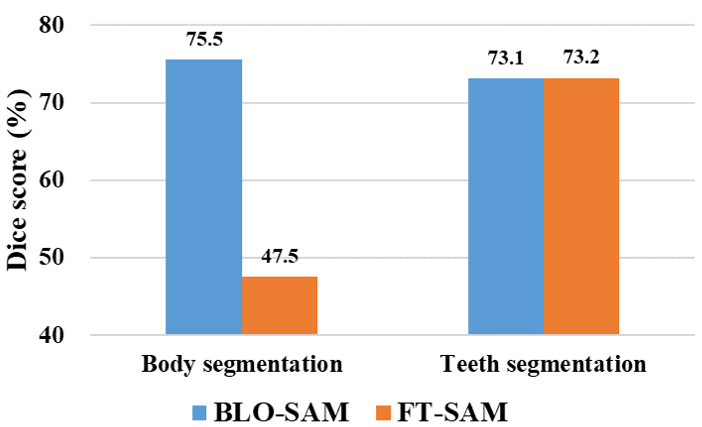}
\vspace{-0.3cm}
\caption{Ablation study of comparing with the full fine-tuning. "FT-SAM" denotes the full finetuning of SAM.}
\label{fig: abla5}
\end{minipage}\quad
\begin{minipage}{.45\linewidth}
\centering
\includegraphics[width=.9\textwidth]{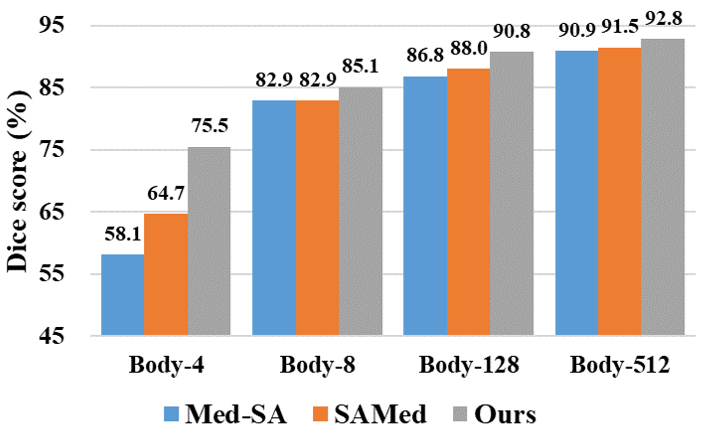}
\vspace{-0.3cm}
\caption{Ablation study of training with different number of examples on body segmentation task.}
\label{fig: abla6}
\end{minipage}
\end{figure}

We conduct experiments to explore what would occur when increasing the number of training examples on the body segmentation task. We increased the number of training sets to 128 and 512, and conducted experiments for Med-SA, SAMed, and BLO-SAM. As shown in Table~\ref{fig: abla6}, After increasing the number of training examples, the performance of BLO-SAM can be further improved, as it achieves the highest dice score of $92.8\%$ among all experiments when training with 512 labeled examples. What's more, when faced with very limited training examples, BLO-SAM also shows a strong capability to overcome the risk of overfitting, which can be demonstrated by the biggest performance gap with Med-SA and SAMed when training with only 4 labeled examples.

\section{Qualitative Results} \label{res: qua}

\begin{figure}[t]
    \centering
    \includegraphics[width=.9\linewidth]{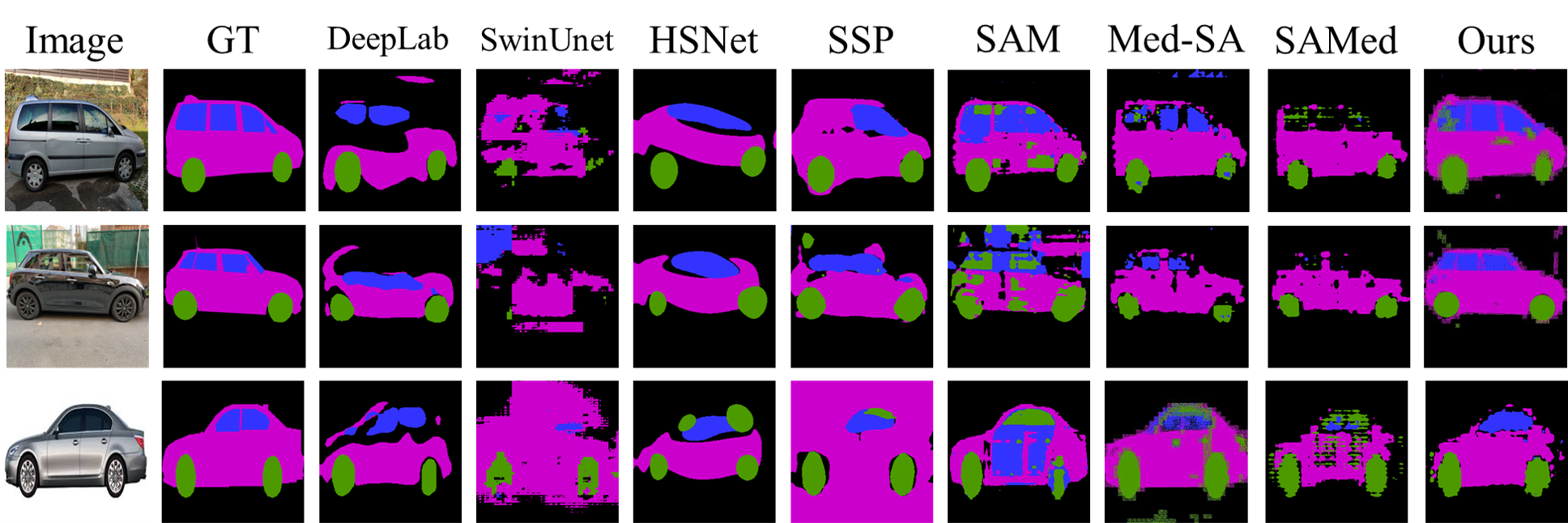}
    \vspace{-0.3cm}
    \caption{Qualitative results on some randomly sampled test examples from the car dataset. 
    }
    \label{fig: qua_car}
\end{figure}

\begin{figure}[t]
    \centering
    \includegraphics[width=.9\linewidth]{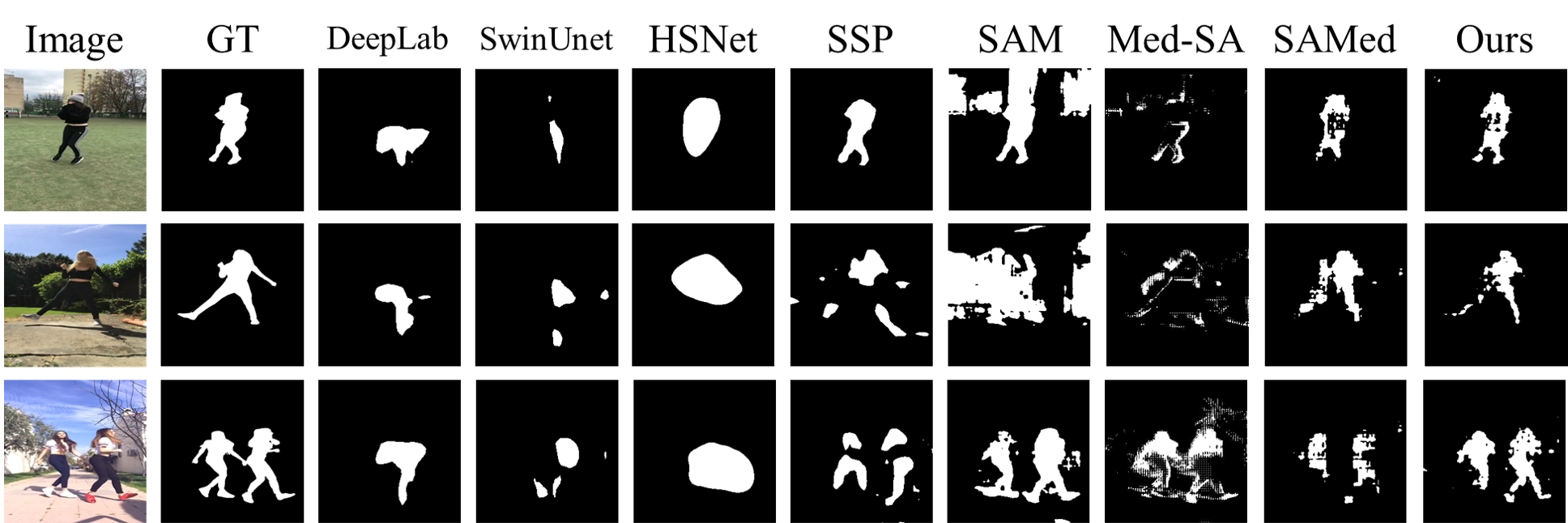}
    \vspace{-0.3cm}
    \caption{Qualitative results on some randomly sampled test examples from the body dataset. 
    }
    \label{fig: qua_body}
\end{figure}

\begin{figure}[t]
    \centering
    \includegraphics[width=.9\linewidth]{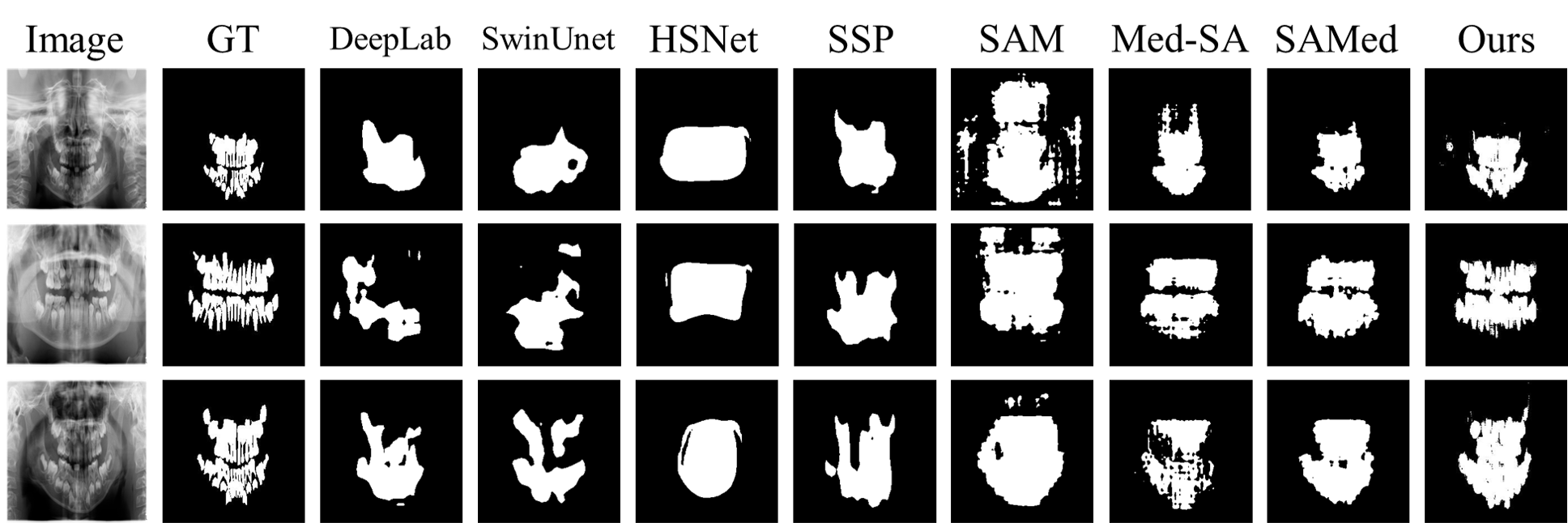}
    \vspace{-0.3cm}
    \caption{Qualitative results on some randomly sampled test examples from the teeth dataset. 
    }
    \label{fig: qua_teeth}
\end{figure}

\begin{figure}[t]
    \centering
    \includegraphics[width=.9\linewidth]{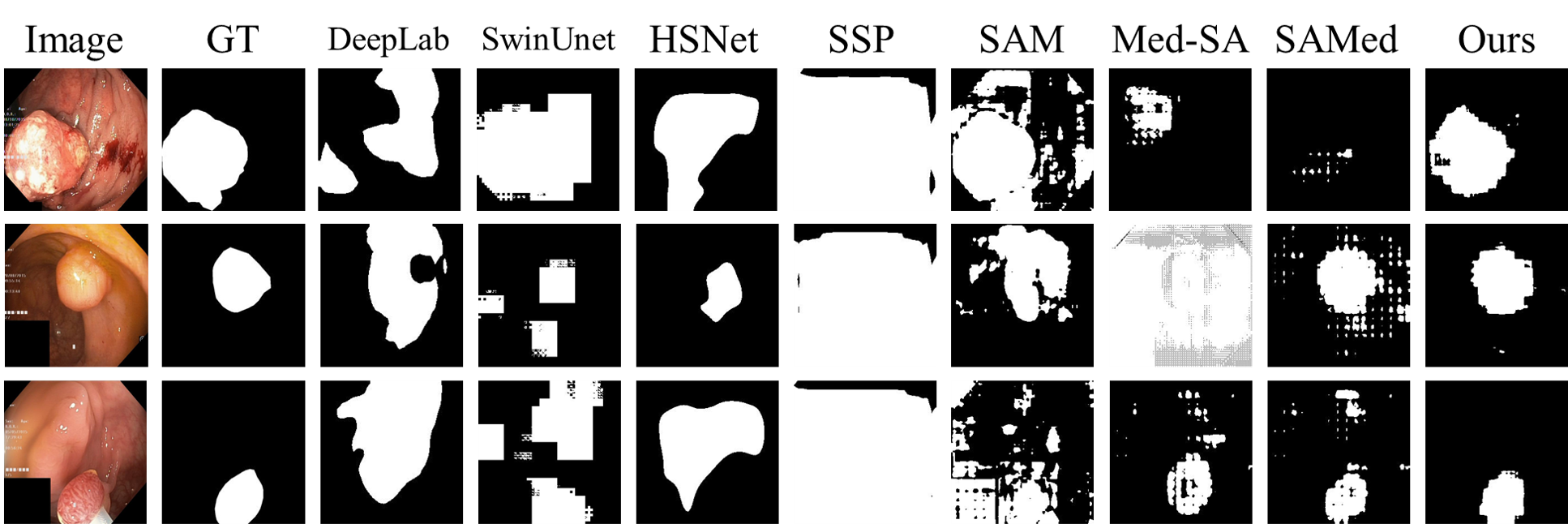}
    \vspace{-0.3cm}
    \caption{Qualitative results on some randomly sampled test examples from the gastrointestinal disease dataset. 
    }
    \label{fig: qua_kvasir}
\end{figure}

\begin{figure}[t]
    \centering
    \includegraphics[width=.9\linewidth]{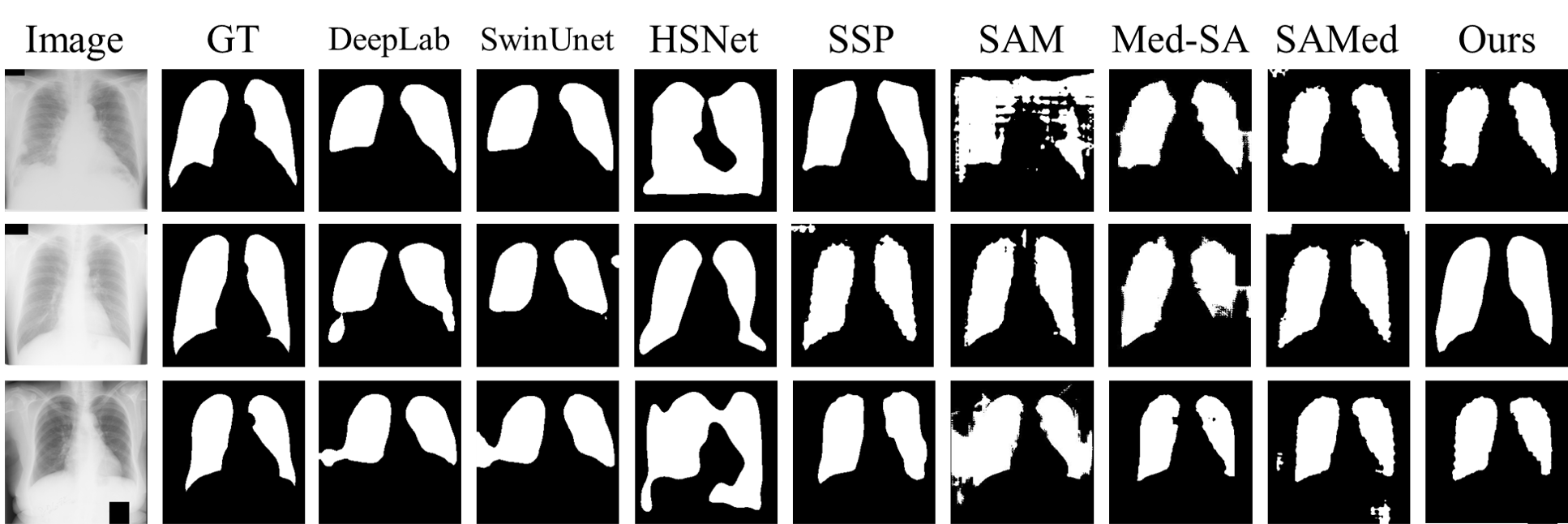}
    \vspace{-0.3cm}
    \caption{Qualitative results on some randomly sampled test examples from the lung dataset.
    }
    \label{fig: qua_lung}
\end{figure}

In Figures~\ref{fig: qua_car}, ~\ref{fig: qua_body}, ~\ref{fig: qua_teeth}, ~\ref{fig: qua_kvasir}, and ~\ref{fig: qua_lung}, we present the segmentation masks generated by different methods. A noteworthy observation is that, across various tasks, the segmentation masks predicted by BLO-SAM consistently exhibit a superior level of detail, demonstrating finer prediction of target components with minimal interference from the background. This is particularly evident when compared to the segmentation masks produced by alternative baselines. The qualitative results underscore the efficacy of BLO-SAM in capturing intricate features while maintaining a cleaner distinction between the foreground and background, emphasizing its robust performance across diverse segmentation tasks.

\end{document}